\pdfoutput=1
\documentclass[10pt,twocolumn]{article} 
\usepackage{simpleConference}
\usepackage{graphicx}
\usepackage{xurl}
\usepackage{url,hyperref}

\begin{document}

\title{Predictive Modeling of Charge Levels for Battery Electric Vehicles using CNN EfficientNet and IGTD Algorithm}

\author{Seongwoo Choi, Chongzhou Fang, David Haddad, Minsung Kim  \\
\\
Institute of Transportation Studies \\
Department of Computer Science \\
University of California, Davis \\
\\
shjchoi, czfang, davhaddad, mngkim @ucdavis.edu  \\
\\
\\
}

\maketitle

\begin{abstract}
Convolutional Neural Networks (CNN) have been a good solution for understanding a vast image dataset. As the increased number of battery-equipped electric vehicles is flourishing globally, there has been much research on understanding which charge levels electric vehicle drivers would choose to charge their vehicles to get to their destination without any prevention. We implemented deep learning approaches to analyze the tabular datasets to understand their state of charge and which charge levels they would choose. In addition, we implemented the Image Generator for Tabular Dataset algorithm to utilize tabular datasets as image datasets to train convolutional neural networks. Also, we integrated other CNN architecture such as EfficientNet to prove that CNN is a great learner for reading information from images that were converted from the tabular dataset, and able to predict charge levels for battery-equipped electric vehicles. We also evaluated several optimization methods to enhance the learning rate of the models and examined further analysis on improving the model architecture.  
\end{abstract}

\section{Introduction}
The rapid evolution of vehicles over the years has greatly improved their efficiency \cite{sulaiman2015review}. With the help of technology, such as hybrid cars and internal combustion engines, transportation has become more efficient \cite{hoyer2008history, santos2021internal}. Data collected by devices through the Internet of Things can be used to improve the system \cite{santos2021internal, leach2020scope}.
Data has been a massive component of modern self-driving technology, which has made it possible to predict the driving patterns of individual drivers \cite{surden2016technological, soni2021design}. It can also help improve the efficiency of the charging infrastructure by analyzing the data collected by the system \cite{soni2021design}.
One of the main advantages of electric cars is their ability to provide single charging~\cite{falahi2013potential, kang2009activity}. Since their energy usage is not as high, it is vital to analyze how much energy they use and how fast they would recharge their batteries.
It is also essential to develop deep neural networks in vehicles to help the utility grid estimate the amount of energy needed each hour.
Due to the increasing number of computer and transportation technology advancements, it is also important to develop deep learning algorithms for electric cars, which will help improve the driving experience of electric vehicle drivers \cite{sanguesa2021review, yang2020efficient, un2017comprehensive, hu2019reinforcement}.

The rapid evolution of transportation has changed various aspects of our society \cite{kuznets1973modern, cano2018batteries}. One of these is the increasing number of people using transportation. As a result, the power distribution challenges are becoming more severe. Electric cars are now considered new concepts of transportation. They are usually powered by either a battery or a hydrogen fuel cell \cite{cano2018batteries}.
Unlike internal combustion engines, electric cars do not emit any pollution when they travel. Instead, they use hydrogen or a battery to power their engines \cite{eberle2010sustainable}. This ensures that they do not hurt the environment by releasing harmful substances.
Even though under most cases electric cars can travel anywhere in a single charge, their batteries still need to be appropriately charged \cite{eberle2010sustainable}, which is why engineers and scientists must continue to study the various aspects of electric vehicle technology \cite{sulaiman2015review}.
Through the development of machine learning and other advanced technologies, researchers were able to identify the most effective ways to recharge electric cars efficiently. One of these is by analyzing the data collected by the electric vehicles and their owners \cite{rezvanizaniani2014review}. 
One of the most critical factors that they need to consider is the appropriate way to predict the charging demand for electric cars \cite{rezvanizaniani2014review, lu2020energy, rahimi2013battery}. In this paper, we present a machine learning method that can help predict the charging levels for electric cars.
The paper presents a deep learning framework that aims to provide the most accurate and efficient method to predict the charging levels for electric cars. We also discuss the various aspects of the system that can be used to improve its system performance.

In this research, we are using the combined tabular dataset of two battery-electric vehicle-related datasets (trip information such as which location the drivers went and what was their vehicles' status in terms of how much full the battery was, what time they started trips while the other dataset contained charging levels that the drivers ended up choosing upon the completion of charging). We will first implement decision trees, kNN, and random forest as baseline models. In addition, we will implement convolutional neural networks and recurrent neural networks for analyzing temporal data. Our contributions in this paper are:
\begin{itemize}
    \item We are the first team to predict battery charge levels on this dataset which contains using user-level information.
    \item We implement and compare several different machine learning methods that can solve this classification problem.
    \item We introduce the Image Generator for Tabular Dataset (IGTD) method to this problem and enable the usage of deep learning models like CNNs in this type of problem.
    \item By integrating IGTD and powerful state-of-the-art CNN model, our prediction accuracy can reach as high as $98\%$.
\end{itemize}

%
\section{Background}
\subsection{Artificial Neural Networks}
A neural network is a set of nodes or units loosely connected to a living brain. The nodes can process and send signals to each other \cite{gurney2018introduction}. The network's output is computed by taking into account the input's number. The connections are called edges, and their threshold prevents a signal from being received if the total amount of information exceeds the threshold \cite{dongare2012introduction}.
A neural network learns by studying examples that contain general information. These examples then form a probability-weighted relationship with the resulting output. The goal of training this system is to determine the difference in output between the expected results.
After noticing that there was a significant increase in the number of people who were buying electric cars, the researchers started looking into the demand for charging stations \cite{zhang2018multi, jahangir2020plug, xu2019utilizing}. They then used the data collected by the National House Hold Survey to develop models that could predict the next trip of the drivers \cite{harris2014empirically, jahangir2020plug}.
The researchers used three standard travel parameters to forecast the trip's duration, start time, and end time\cite{harris2014empirically}. They noted that the course of the trip could improve the accuracy of their predictions. They also used various models, such as random forest, KNN, and deep ANN, to analyze the data \cite{xu2019utilizing}.
The researchers found that the travel behavior of car owners can help predict the daily charging demand of an EV. They were able to do so by analyzing the various trip parameters of an EV using a statistical model \cite{de2015artificial}. Their study showed that a machine learning approach based on the NHTS survey data was able to improve the accuracy of their predictions.
The researchers concluded that having multiple parameters can boost the accuracy of the models. They also suggested that the training of artificial neural networks should consider the baseline models.
The researchers built their models on various types of neural networks, such as multi-layer perceptrons \cite{jahangir2020plug} and recurrent neural networks. They were able to predict the destination of a taxi based on its beginning sequence.
A multi-layer perceptron is a neural network that uses multiple layers to process input \cite{zhang2018multi}. It takes into account the input's fixed-size vectors and maps them to higher levels of representations \cite{de2015artificial}.
The researchers' model was able to perform an almost fully automated task by identifying the destination of a taxi based on its initial sequence. It used a recurrent neural network to encode the prefix and several embeddings to generate its output \cite{zhang2018multi}.
For their research, the researchers considered the cluster as a parameter of the network. They then trained the system to learn to accept either random or mean-shift clusters.
The increasing number of plug-in electric cars has led to the need for more accurate and timely predictions about their potential demand \cite{mukherjee2014review, darabi2012impact, kong2016charging}. The researchers noted that this is important because of the high uncertainty in the drivers' behavior \cite{jahangir2019charging}. Previous research reported that predicting the demand for electric cars based on traditional vehicles can cause issues since this could lead to bias in the charging pattern and power estimation \cite{kong2016charging,shuai2016charging}.
The researchers then applied a feed forward and recurrent neural network approach to forecasting the trip's PEV-TB \cite{jahangir2019charging}. They also used historical data such as the arrival time and the length of the trip. They were able to reduce the financial loss of the taxi aggregators by around 16 dollars per year.
The researchers then trained a rough artificial neural network to forecast the travel behavior of plug-in electric vehicles \cite{zhang2018multi, jahangir2020plug, xu2019utilizing}. It improved its accuracy by using rough neurons in the recurrent network.
The researchers were able to use the data collected by the GPS-equipped probes to estimate the travel time of a given target segment\cite{zhan2013urban, herrera2010evaluation}. They considered the spatial-temporal relevance of the data to the target segment's travel time \cite{zhang2018multi, jahangir2020plug, xu2019utilizing}. For instance, if a target segment has a history of experiencing significant travel times, its travel time might be related to the traffic conditions in nearby areas.
After training the system, the researchers applied a network clustering algorithm to analyze the data and perform a travel time distribution analysis. They then used an artificial neural network to study the complex spatial-temporal relevance of the data \cite{jahangir2019charging}.
The researchers then clustered relevant segments into a group to build ANN models that can improve the accuracy of their predictions. They used various functions such as activation functions and hyperbolic tangents.

%
\subsection{Convolutional Neural Networks}
Convolutional Neural Networks (CNN) are a set of deep learning models which can take in an input image, assign importance (learnable weights and biases) to various aspects/objects in the image and be able to differentiate one from the other \cite{ghildiyal2020age}. CNNs have been successfully used in various applications, such as image recognition and video analysis, natural language processing, and speech recognition \cite{yamashita2018convolutional}. They are inspired by the visual neuroscience. They have various features that can be used to analyze and interpret natural signals, such as connections in receptive field~\cite{zhu2021converting}. These features allow CNNs to perform analysis on data with temporal or spatial dependencies between components. For instance, in imaging, they can detect the spatial arrangement of the pixels in an image. Also, they can identify low-level features in the image\cite{zhu2021converting}. 

Previously, several models received great attention and have been widely deployed in image-related tasks, e.g. ResNet~\cite{he2016deep}, ImageNet~\cite{krizhevsky2012imagenet}, etc. In 2019, EfficientNet~\cite{tan2019efficientnet} was proposed, which greatly improves the efficiency of CNNs in the literature, and at the same time reach state-of-the-art accuracy. The authors study the scaling problem of CNNs and craft a way to properly scale baseline CNN model with given resources to obtain optimal accuracy. In this work, we will utilize EfficientNet-B0~\cite{tan2019efficientnet} to enhance our classification performance.

The trajectory data is transformed into two-dimensional trajectory images, which ensures the temporal and spatial relationships between the trajectory data \cite{tan2019efficientnet, liu2012understanding}. The trajectory images are converted from spatial domain to frequency domain to be able to reduce the noises of the image and alleviate the sparsity using Fast Fourier Transform \cite{zhang2018multi}. The traditional taxi prediction methods model the taxi trajectory as a sequence of spatial points. It cannot represent two-dimensional spatial relationships between trajectory points. Therefore, many methods transform the taxi GPS trajectory into a two-dimensional image, and express the spatial correlations by trajectory image \cite{kong2016charging, jahangir2019charging}. The Pooling layer is responsible for reducing the spatial size of the Convoluted Feature.

%

\subsection{Recurrent Neural Networks}
A recurrent neural network (RNN) is an artificial neural network that connects nodes in a temporal sequence \cite{li2017diffusion}. It can exhibit temporal dynamic behavior. RNNs are similar to feedforward neural networks, which can handle variable-length sequences \cite{abiodun2018state}.
These networks are commonly used for various tasks such as speech recognition and unsegmented handwriting recognition. Recurrent neural networks are capable of running arbitrary programs to process data \cite{TEALAB2018334}.
The term recurrent neural network is sometimes used to refer to a network with an infinite impulse response. A recurrent network can be replaced with a feedforward neural network\cite{li2017diffusion}.
Recurrent networks can also have additional stored states, which the network can control. Other networks or graphs can replace these states with feedback loops or time delays. These states are called gated memory and are part of long-term memory networks.
Since the task is time series analysis using temporal information\cite{li2017diffusion, TEALAB2018334}, it is necessary to attempt recurrent neural network modeling to understand the temporal data analysis. Long Short-Term Memory was used to solve the Vanishing Gradient problem of General RNN and then tested it with a simple model (epochs = 10). 
Since the dataset has values that consider temporal data, such as the distance each trip traveled in a specific time period and how long they traveled, eventually choosing which charge levels, recurrent neural networks would be an excellent method to understand the dataset more.

%




%

\subsection{Battery Electric Vehicles}
Instead of using internal combustion engines, BEVs (batteries electric vehicles) use electric motors and controllers to power their wheels. These vehicles are not limited to cars or motorcycles. They can also be used in various types of equipment such as boats, bicycles, buses, and railcars \cite{rahimi2013battery}.
A battery electric vehicle is a type of vehicle that uses chemical energy stored in a battery pack. It can be categorized into four types: pure electric, all-electric, hybrid, and pure electric \cite{zeng2019commercialization, li2019comprehensive}. These vehicles have no secondary sources of propulsion, such as internal combustion engines or hydrogen fuel cells.

%

\subsection{Dataset Information}
The dataset that is used for the purpose of this research was given by the Institute of Transportation Studies at the University of California, Davis. The dataset includes all of trip information with day of the trip, the hour each trip started and ended, the state of battery such as percentage of battery charged in the beginning of each trip and how much percentage the battery drained after each trip. The day name from Monday to Sunday and day type such as weekday and weekend. Which model each driver was driving and which car maker they were driving. How much distance each trip traveled from which origin location and destination of each trip. The dataset also includes year, month, day, and how much time battery was charging, which could illustrate whether each driver was using slow charging or fast charging. 

Since collecting such vast information is not trivial and predicting which charge level that a driver would eventually choose to charge battery electric vehicles is important to understand where the charging infrastructure should be built throughout the nation, it became a significant task to understand driving patterns of drivers. Understanding driving patterns of drivers would be important because every driver has a different style of driving a car and it may drain battery due to their usages. In addition, limited resources to build charging infrastructures have become another reason to conduct such research. Allocating limited resources appropriately is a key factor to make charging infrastructures prevalent and ubiquitous so that more drivers can rely on their battery equipped electric vehicles and do not prevent them from driving their vehicles for long trips. Providing a reliable deep learning algorithm and predictive analysis is the main task of the research.  

\begin{figure}[h]
\centering
\includegraphics[width=\linewidth]{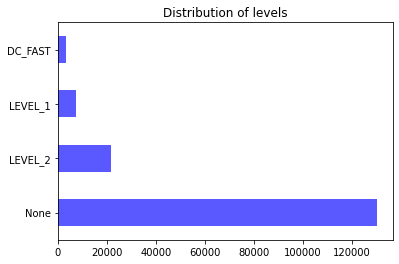}
\caption{The Original Dataset from Battery Electric Vehicles for Each Charge Levels}
\end{figure}

The dataset we have given as shown above clearly states that there are imbalanced data in the dataset. Having more 'None' data is usual because not every driver would charge their vehicles in every single moment after each trip. Some trips could be short such as traveling only ten minutes to get to their closest destination, or there might be no charging infrastructure around them. Just because driving a battery equipped vehicle does not mean that there should be some sort of charging event after each trip. What makes this research not trivial is that there should be an algorithm to understand why a driver would eventually choose none charging, or slow charging, or fast charging. Is it because of the trip was too short or was it because there was no charging station around them? Some data show that they traveled a long distance, but did not choose to charge their vehicles during the nighttime. Assuming the driver had a charging station built in at their homes, why was there no charging event whatsoever? The research attempts to answer such questions and challenges to provide the best predictive model for electric vehicles so that more people end up choosing the electric vehicles for their transportation method and bring sustainable energy to our surroundings. 

%
\section{Limitations}
Though there are a few previous works focusing on predicting state of charge (SoC) of battery~\cite{how2019state}, current machine learning algorithms deployed in electric vehicle (EV) does not pay much attention to the charging level, which plays an important role in estimating the charging demand for these vehicles. As more EV owners choose EVs for their primary transportation method, it is important to accurately predict the charging level because EVs' optimal charging and scheduling under dynamic pricing schemes will dramatically reduce carbon emission while traveling. Our project aims at comparing different approaches in deep neural networks to fully understand which architecture performs the best in terms of creating an optimal deep learning solution for battery-equipped vehicles that considers the charging problem from the utility (Power distribution system) point of view. We want to know how much electricity we should provide for our fleet. The novel aspect of our research project is that we dive into exploring the critical areas of modern-day driving technology, and we want to know that the utility grid can provide the energy needed by the fleet at the right charging location. We will propose new models on a new real-world dataset to solve this prediction problem and use features different from previous SoC prediction studies~\cite{ng2020predicting}.

%
\section{Methodology}
In this paper, we will implement the deep neural network approaches such as EfficientNet~\cite{tan2019efficientnet}, or similar architectures for a higher accuracy as a primary use case aiding in driving battery electric vehicles by analyzing trip summary data and then predict whether they will charge their car at the end of the trip based on the information at the start point of trip, and if they charge, what charge level they will use. We will (1) predict whether the EV will charge their car at the end of the trip based on the information at the start point of trip, and (2) if they charge, what charge level they will use. We will mainly utilize the power of deep neural networks. We will be using convolutional neural networks with and without jittering, and compare the performance of those results in predicting charging levels of EV batteries. Essentially, we want to understand the relationship between the charging behavior and the information at the beginning of the trip. In addition, we are suggesting a novel way to formulate the problem as a classification problem, where there are 4 different types of results (0-3) representing different charging levels:
\begin{enumerate}
    \item 0 - charging level 1;
    \item 1 - charging level 2;
    \item 2 - DCFAST;
    \item 3 - None charging.
\end{enumerate}
In addition, predicting which charging method the driver will use is a hard challenge because understanding the driving pattern of each trip is complicated. To understand the driving pattern may require more advanced algorithm for further analysis in the dataset. However, if we could predict the charge levels through the research, then understanding their driving patterns would be more handy for further research projects. 

Previous works use relatively simple models like SVM, regression, etc.~\cite{zahid2018state, ren2018remaining, hussein2014kalman, dawson2018data}. In order to have the model to understand all of those input features, we should also consider that we can first utilize the multi-layer perceptron and then move on to convolutional neural networks. We will implement a simple one-layer perceptron or logistic regression for comparison for the baseline architecture and how the model improves when training on other architecures. We will also add jittering~\cite{nourani2018hybrid} to the MLP model ~\cite{7535559} and the CNN model during the training to see if adding jitter data would improve the prediction. For using the CNN model, since it is used mostly in images, we will consider converting embedding data into images and then train the model based on those image data ~\cite{yamashita2018convolutional}.

Evaluation on real-world EV data sets is not sufficient in literature~\cite{hussein2014kalman,chemali2018state}. Also, they heavily rely on low-level circuit metrics~\cite{sahinoglu2017battery,tong2016battery} which may not easy to accurately collect in real-world scenarios. In this project, we will be using a novel dataset with data points collected from real EV batteries and travel information at user level during driving. Instead of including only electrical features, this dataset emphasize more on user-level metrics. We are considering using the following features from the dataset as input features:
\begin{enumerate}
    \item Start Time - Indicates when the driver started the engine, i.e. when he or she started the trip
    \item SoC Start - Percentage of the battery when the trip started
    \item Day Type - Weekday or weekend
    \item Day Name - Monday to Sunday
    \item Vehicle Model - Model names, e.g. Tesla, Toyota, Chevrolet, and Nissan
    \item Holiday - whether it was during the holiday because people usually travel during the holiday
    \item Origin - Other, home, work
    \item Day of the year - 1-365 (normalize the data)
    \item Seasons - Spring (0), Summer (1), Autumn (2), Winter (3) [normalize]
    \item Destination - home, work and others.
\end{enumerate}

We will further explain how to utilize Recurrent Neural Networks, and Convolutional Neural Networks for analyzing electric vehicle datasets in the next sections.

%
\subsection{RNN Approach}
We first selected input features and created a new dataframe. For each label in the dataset, we have converted them into embedding using LabelEncoder and fit\_transform for normalize the data that we were interested in analyzing. We then applied one-hot encoding for target variables that were categorical. MultiLabelBinarizer could converted and fit those target labels. With four classes that we got for target variables, we then analyzed the shapes of x and y values that we divided for training purpose. With train-test split, we divided training and testing sets with test\_size of 0.3. 

We built a Long Short-Term Memory sequential model with 50 LSTM layers, and Recified Linear Unit (ReLU) activation function, and had an input shape of (27, 1). We utilized dropout of 0.1 and dense layers of 5 and 1. When compiling, we used adam optimizer and Mean Squared Error (MSE) loss, and accuracy metrics. 

For training the model, we ran 15 epochs with batch size of 30 and used early stopping on validation loss. However, Recurrent Neural Networks (RNN) may seem valid at predicting the charge levels, they do not predict every single fleet in the tabular dataset. RNNs do not generalize well when there are more than one vehicle model in the dataset, and there were several research conducted to understand the overall fleet prediction based on their temporal data, but the RNNs did not show any great performance. In addition, that is not the main focus of the research, so we have determined to move forward and utilized performance we could get from the convolutional neural networks (CNN) to overcome the barriers we faced when we were using RNN. We will demonstrate how we could use convolutional neural networks to understand the whole tabular dataset. 

%
\subsection{CNN Approach}
Adapting and using models well developed in a different field can sometimes bring surprisingly superior performance. For example, transformers~\cite{vaswani2017attention} was previously widely utilized in natural language processing, while CNNs dominated image-related tasks a few years ago. However, research works like~\cite{dosovitskiy2020image} reveal that attention mechanism existing in transformers can also be applied to image-related tasks and the resulting visual transformers achieve better performance than CNNs.

In this project, we want to utilize advanced CNN models like EfficientNet-B0~\cite{tan2019efficientnet} to perform classification tasks, instead of only using basic machine learning models~\cite{goodfellow2016deep}. We need to first perform preprocessing of our data to transfer our tabular data to images. The resulting steps can possibly better capture correlations of data. Next, we will be utilizing off-the-shelf powerful CNN models to perform our classification tasks.

Since convolutional neural networks are generally used in images, it is necessary to convert tabular datasets to images. It is a challenging task to fill in the gap between tabular data and 2D images, since there are multiple aspects to consider. For such approach, we have utilized the Image Generator for Tabular Dataset (IGTD) Algorithm to convert tabular datasets into 2d images in both grayscale and color using Euclidean distance \cite{breu1995linear} and Manhattan-Pearson \cite{chang2009compute} calculation \cite{zhu2021converting}. 

\subsubsection{IGTD Algorithm}
In \cite{zhu2021converting}, the researchers have proven an original algorithm to convert a tabular dataset into embedding and then convert the embedding dataset into grayscale images by calculating and ranking pairwise distances of features. The researchers presented a novel method to transform data into images for deep learning analysis using CNNs. The method generates an image by assigning a feature to a pixel in the data. The intensity of the feature in the image is then reflected in the corresponding data sample \cite{zhu2021converting}. The goal of the algorithm is to find an optimal assignment of features to pixels based on the distance between the features and the pixels. The distances between the features and the pixels are computed using the coordinates of the image \cite{zhu2021converting}. The goal of this algorithm is to reduce the difference between the two top rankings by assigning similar features to the neighboring pixels and contrasting features to the far-off ones. Through an iterative process, the algorithm selects the features that are most suitable for reducing the difference between the rankings. Third, the number of dimensions, size, and shape of the images can be flexibly chosen. \cite{zhu2021converting}.

\subsubsection{Euclidean Distance}

\textbf{X} denote an M by N tabular data matrix to be transformed into images. Each row of \textbf{X} is a sample and each column is a feature from the tabular dataset. Let $x_i$,:, $x_j$ denote the ith row, the jth column, and the element in the ith row and jth column, respectively. The algorithm uses euclidean distance to create an $N_i$ by $N_j$ image (i.e. a 2-D array), where $N_i$×$N_j$=N. The pairwise distances between features are calculated according to a distance measure, such as the Euclidean distance. The pairwise distances are then ranked ascendingly, so that small distances are given small ranks while large distances are given large ranks. An N by N rank matrix denoted by \textbf{R} is formed, in which $r_{i,j}$ at the ith row and jth column of \textbf{R} is the rank value of the distance between the ith and jth features. Details regarding the data will be presented in the next section. Distances between genes are measured by the Euclidean distance based on their expression values. The larger the rank is and the darker the corresponding point is in the plot.

\textbf{Definition}: Distance between two points \newline

\begin{center} 
    ($\mathbf {p,q}$)= $\sqrt{\sum \limits_{i=1}^n (q_i-p_i)^2}$ 
\end{center}

\bigskip

p, q	=	two points in Euclidean n-space. \newline
$q_i$, $p_i$	=	Euclidean vectors, starting from the origin of the space (initial point) 
n	=	n-space

\begin{figure}[h]
\centering
\includegraphics[width=\linewidth]{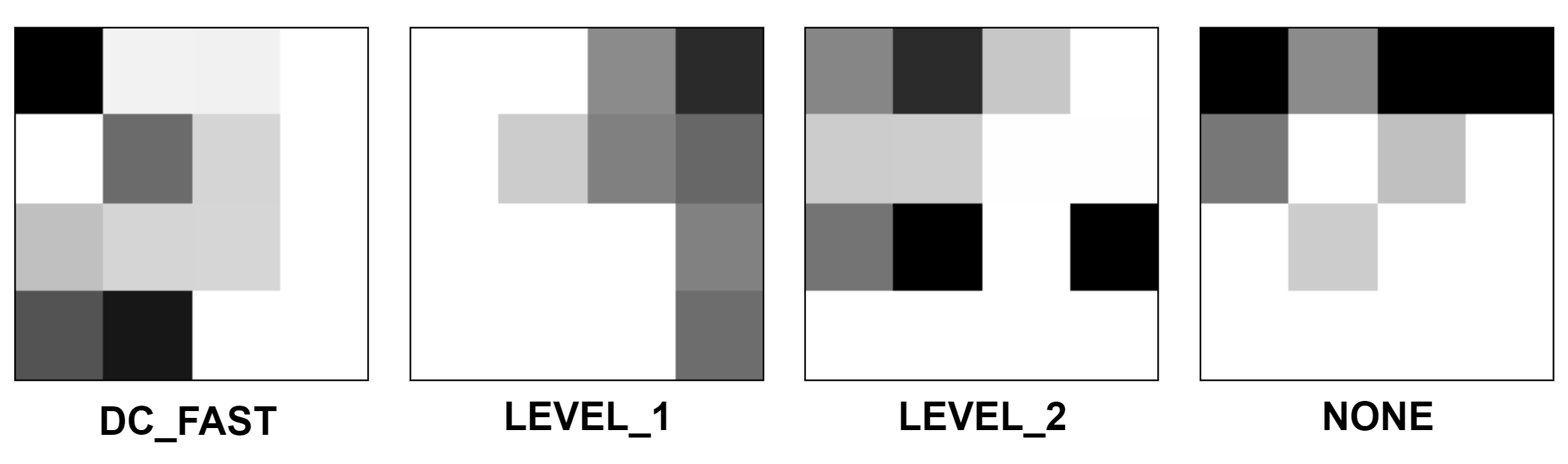}
\end{figure}

[more explanation here]

\subsubsection{Manhattan-Pearson Distance}

Manhattan distance can also be used instead of the Euclidean distance. Different functions can be utilized to measure the deviation between the feature distance ranking and the pixel distance ranking. The different functions can show distinct aspects of the data. An example, the difference between absolute difference function the squared difference function is that one puts larger weights on elements with large differences \cite{chang2009compute}. 

\textbf{Definition}: The distance between two points measured along axes at right angles. In a plane with p1 at (x1, y1) and p2 at (x2, y2), it is |x1 - x2| + |y1 - y2|.

\begin{figure}[h]
\centering
\includegraphics[width=\linewidth]{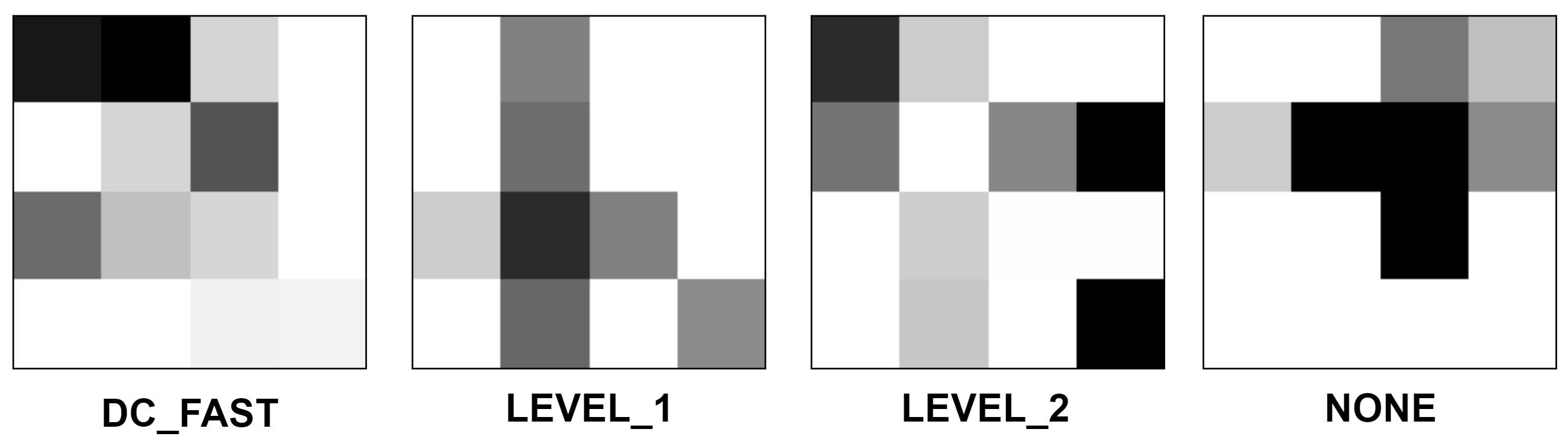}
\end{figure}

Each feature in a tabular data file needs to be assigned a pixel position in the image to make it easier to transform it into images. A simple way to do this is to divide the feature distance rank matrix R into the ith and ith pixel. However, when we compare the two matrices, we can see that the ith and the ith row have different diferences \cite{zhu2021converting}.
The error function is used to measure the diference between the two symmetric matrices. It can be used to compute the absolute or square diference between ri and j. There are also various options when it comes to calculating the diference between these two matrices.
The task of assigning features to a suitable pixel position in an image is now done so that they are close enough to each other that it makes sense to err(R, Q) less. This step also involves making sure that the order of features along the columns and rows in R is synchronized.
The concept of a feature reordering is to swap the positions of two features in a sequence. In order to reduce the error, we can search for suitable feature swaps. This algorithm is based on the idea of the IGTD algorithm \cite{zhu2021converting}.

In \cite{zhu2021converting}, the researchers introduced a novel method to transform tabular data into images for deep learning analysis. The method generates an image for each data sample by assigning each feature a pixel in the image. The goal of this algorithm is to find an optimal assignment of features to the selected pixel. The distances between the features and the assigned pixel are computed based on their coordinates, and the goal of this method is to minimize the diference between the two images by assigning dissimilar and similar features to each neighboring pixel. In order to achieve this, an iterative process is performed to swap the assignments of two features. The algorithm identifies the feature that it considers most likely to perform well in the swap, and it then seeks to find a feature that it can reduce the diference between the two images.

\subsubsection{Convolutional Neural Network with EfficientNet}
The EfficientNet~\cite{tan2019efficientnet} framework is a semi-supervised network architecture that uniformly scales the dimensions of a network's depth, width, and resolution dimensions. Targeting certain resource constraints, EfficientNet can use a compound scaling method to intelligently scale the architecture of a baseline model and achieve optimal performance with minimal computation overhead. By integrating EfficientNet into our architecture, we are able to utilize its power in image-related tasks and achieve high accuracy performance.

\begin{figure}[h]
\centering
\includegraphics[width=\linewidth]{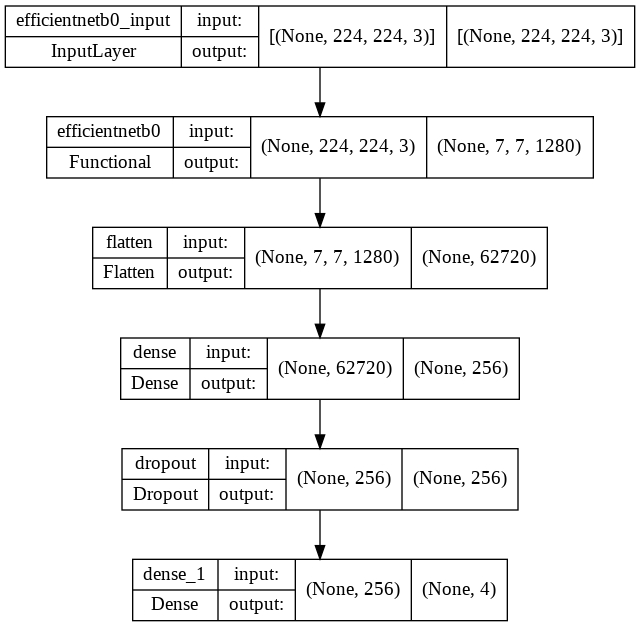}
\caption{Model Architecture for our architecture.}\label{FigArchitecture}
\end{figure}

In our model implementation, we utilize EfficientNet-B0, which is the most lightweight model in the EfficientNet family~\cite{tan2019efficientnet}. After passing EfficientNet-B0, the data will flow through fully-connected and dropout~\cite{srivastava2014dropout} layers for further processing. The complete model architecture is shown in Figure~\ref{FigArchitecture}.

As shown in Figure~\ref{FigArchitecture}, the inputs to our model are image data that are converted from the original tabular dataset. The pre-trained EfficientNet-B0 model constitutes the main complexity of our architecture. After the processing of several fully-connected layers and dropout layer, classification results of the four classes are generated.

%

\bigskip

%
\section{Results}
In this section, we will explain our findings on baseline machine learning models such as random forest, k nearest neighbors, and decision tree. After that, we will go over our findings in recurrent neural networks, and convolutional neural networks. Especially, we will dive deep into convolutional neural networks to illustrate our methodologies using other model architecture such as EfficientNet and compare it with the baseline convolutional neural network model we built to see if there was a significant improvement over the baseline model. 
%
\subsection{Baselines for Predictive Modeling}
\subsubsection{Random Forest}
A random forest (RF) is a learning method that involves constructing a large number of decision trees at a time. This method is used for various tasks such as classification and regression. The output of the forest is then used to select the classes that should be included in the classification task.
The random decision forests are more accurate than the trained decision trees when it comes to making decisions. However, their accuracy is lower than that of gradient boosted trees due to their data characteristics. 

For this project, we have chosen those input features that were previously addressed to analyze the dataset. We have converted categorical datasets into embedding using one-hot encoding and we have utilized 10 fold cross validation accuracy. We have reduced the number of data by sampling down to 9000 for level 2 charge level and none charge level due to imbalanced data. To match the number of data available for the rest of the data we have for each charge level, we had to reduce the number of data that exceeded 9000 or more. To create sampling randomly, we have utilized random state method to sample data from those data that exceeded 9000 or more. 

We have a result of 67$\%$ accuracy. 
\begin{figure}[h]
\centering
\includegraphics[width=\linewidth]{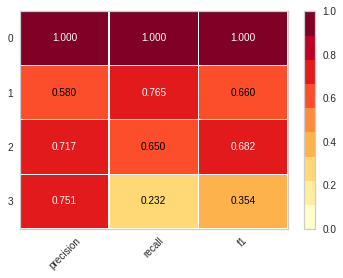}
\caption{Random Forest Result in Classifier Classification Report}
\end{figure}
%
\subsubsection{K Nearest Neighbors}
The k-nearest neighbors (KNN) algorithm is a non-parametric method for learning statistics. It can be used for both classification and regression. The output depends on whether the method is used for these two applications.
The output of a class is a representation of an object's membership in its class. An object is classified by its neighbors, with the most common class being its k nearest neighbor.
The output of k-NN regression is the property value of an object. It is the average of the values of the nearest neighbors.
The k-nearest neighbors algorithm is a type of classification that only approximates the function locally. Since its distance is used to determine its classification, it can improve its accuracy by taking into account the varying scales of its features.
For both classification and regression, a technique known as weighted relations can be used to determine the contribution of the nearest neighbors to the average.
The k-nearest neighbors algorithm takes into account the class or property value of an object's nearest neighbor. This is a set of objects that are used for its training. We have achieved 62.5$\%$ accuracy. 
\begin{figure}[h]
\centering
\includegraphics[width=\linewidth]{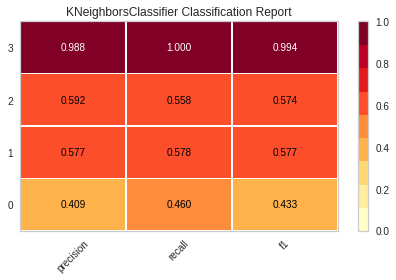}
\caption{KNN Result in Classifier Classification Report}
\end{figure}

%
\subsubsection{Decision Tree}
\begin{figure}
\centering
\includegraphics[width=\linewidth]{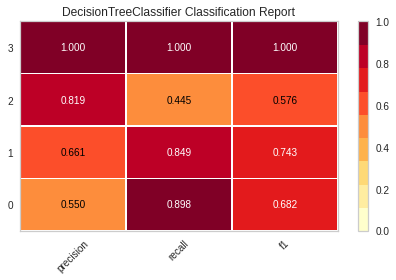}
\caption{Decision Tree Result in Classifier Classification Report}
\end{figure}

A decision tree is a tool that supports decision making by giving a tree-like model of decisions that can be predicted with varying consequences. It can also display an algorithm that only has conditional control statements.
A decision tree (DT) is a type of research that helps identify a strategy that is most likely to yield a goal. It can also be used in machine learning to learn more about a given strategy. We have achieved 73.8$\%$ accuracy.

%

\subsection{Recurrent Neural Networks}
We first selected input features and created a new dataframe. For each label in the dataset, we have converted them into embedding using LabelEncoder and fit\_transform for normalize the data that we were interested in analyzing. We then applied one-hot-encoding for target variables that were catergorical. MultiLabelBinarizer could converted and fit those target labels. With four classes that we got for target variables, we then analyzed the shapes of x and y values that we divided for training purpose. With train-test split, we divided training and testing sets with test\_size of 0.3. 

We built a Long Short-Term Memory sequential model with 50 LSTM layers, and Recified Linear Unit (LeRU) activation function, and had an input shape of (27, 1). We utilized dropout of 0.1 and dense layers of 5 and 1. When compiling, we used adam optimizer and Mean Squared Error (MSE) loss, and accuracy metrics. 

For training the model, we ran 15 epochs with batch size of 30 and used early stopping on validation loss. We have achieved 75$\%$ of accuracy. 

\begin{figure}[h]
\centering
\includegraphics[width=8cm]{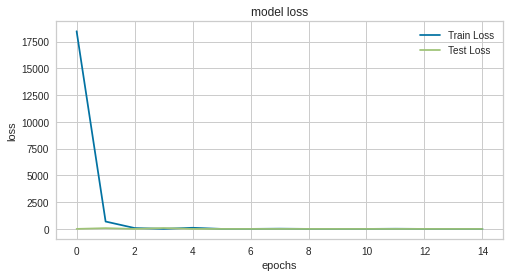}
\caption{RNN: Training and Validation Loss}
\end{figure}

Although this looks very promising in terms of predicting, the confusion matrix result shows otherwise. The RNN model we developed and implemented could only predict one vehicle model at the time. Our goal is to understand the whole dataset and then understand the driving patterns, then predict which charge levels that a driver would choose. Determining such computation would require other model architecture and it would be better to have more data input that generalize well when there is a new set of data for future usage. 
We will switch our focus to the convolutional neural network baseline model architecture and how it showed a better result than that of recurrent neural network we have shown.

\bigskip

%
\subsection{Convolutional Neural Networks}
After converting a tabular dataset into images using IGTD algorithm, we classified those images into four classes: DC\_Fast, Level 1, Level 2, and None. With such classification, we expected the model to understand and extract features from those converted images that were in four by four matrix. With labels ignored, we have divided the image dataset for training and validation purposes. Training dataset consisted of 80\% of original image dataset and validation was 20\%. 

At first, the base convolutional neural network (CNN) model was overfitting on the training data and achieved and incorrect accuracy of approximately 97\%. To tackle the overfitting model, we added 20\% dropout layers after each pooling layer and added L1 and L2 regularizers in each convolution layer. We've trained the model on various batch sizes starting with 32, 64, 128 and 256. There were minimal changes in accuracy and thus leaving the \textit{batch size 32} for the rest of the training. \newline Different optimizers we used to test between Adam and Stochastic Gradient Descent (SGD); resulting in \textit{Adam} as the best performing optimizer. \newline Loss Function used was \textit{Sparse Categorical Crossentropy} and trained on 3 ,5 ,10 and finally 15 epochs. \textit{15 epochs} was used for the rest of the training for RNN and other CNN model architecture. 

\begin{figure}[h]
\centering
\includegraphics[width=8cm]{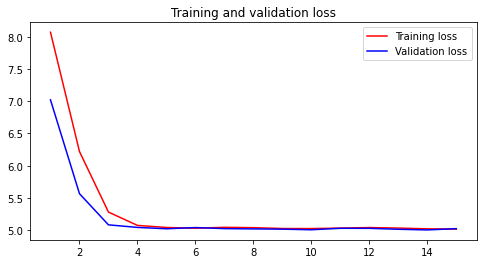}
\caption{CNN: Training and Validation Loss}
\end{figure}

\begin{figure}[h]
\includegraphics[width=8cm]{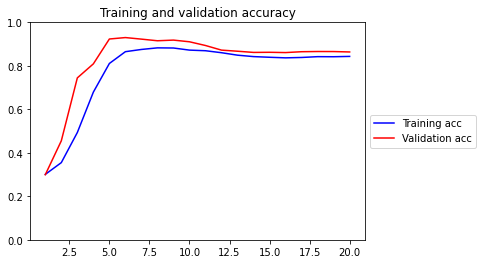}
\caption{CNN: Training and Validation Accuracy}

\includegraphics[width=8cm]{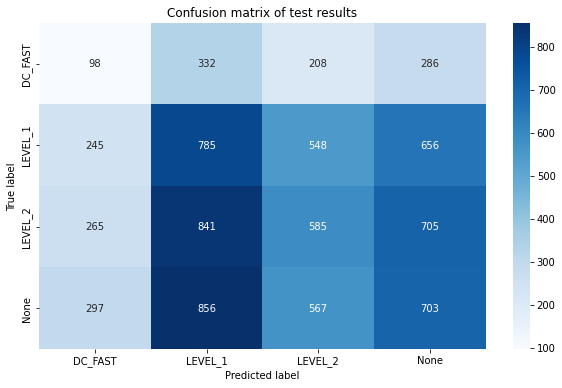}
\caption{CNN Baseline Confusion Matrix}
\end{figure}

The training and testing has been duplicated for the Pearson Manhattan generated images as well. Accuracy was approximately the same as Euclidean distance image which led us to believe that the CNN model could be tweaked a bit more to improve the accuracy of the images. Optimized parameters are the following: Batch Size 32, Adam optimizer, Sparse Categorical Crossentropy loss function, and 15 epochs which led to an accuracy of \textbf{79.95\%}

Researchers are now using CNN, a class of deep learning networks, for various applications such as image classification. Unlike traditional methods, CNN does not require preprocessing the images. It can be used for various tasks such as image segmentation and batch processing. CNN's features are a huge breakthrough in image classification as it eliminates the need for preprocessing.

From the results above, the convolutional neural network baseline model architecture was able to capture some of the image features that were extracted and learned during the model training. However, the CNN model could not capture a lot of details for DC Fast, and Level 1. While not overfitting the model, we wanted to expand from where we have from the baseline CNN model, we have adopted and implemented one of the most updated and advanced method in the convolutional neural networks, EfficientNet, to overcome some of the challenges. The next section explains our findings and evaluation on EfficientNet implementation to tackle this challenge.

\subsection{EfficientNet Result}
\begin{figure}[h]
    \centering
    \includegraphics[width=8cm]{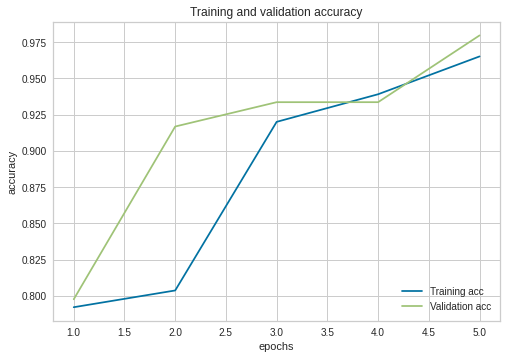}
    \caption{EfficientNet: Training and validation accuracy.}
    \label{FigENetTrainingValidation}

    \centering
    \includegraphics[width=8cm]{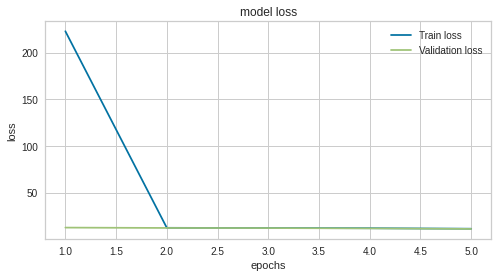}
    \caption{EfficientNet: Training and validation loss.}
    \label{FigENetLoss}
\end{figure}

We integrate EfficientNet-B0~\cite{tan2019efficientnet} and train the model on the processed image dataset. We run the training for 5 epochs, using BCE loss and Adam optimizer with batch size 35. Results regarding our training process are shown in Figure~\ref{FigENetTrainingValidation} and Figure~\ref{FigENetLoss}. 

The accuracy result of our model with EfficientNet is $97.97\%$, which is the highest compared with other methods. We conclude that this is because: (1) EfficientNet-B0 we are using is a complicated and powerful model, which can effectively capture patterns existing in images; (2) Integration of EfficientNet benefits from transfer learning~\cite{pan2009survey}, since the model we integrate is pre-trained on the large ImageNet dataset~\cite{deng2009imagenet} and is capable of handling different classification tasks.

%
\subsection{Summary Comparison}
As a summary, the accuracy comparison of different models are shown in Table~\ref{TableComparison}. We can see that by utilizing novel approaches to convert tabular data to images, we are able to utilize state-of-the-art CNNs and greatly boost the performance.

\begin{table}[h]
    \centering
    \begin{tabular}{cc}
    \hline
        Method & Accuracy\\ \hline
        Random Forest & $67\%$ \\
        K-Nearest Neighbors & $62.5\%$\\
        Decision Tree & $73.8\%$\\
        RNN & $75.00\%$ \\
        Basic CNN & $86.52\%$\\
        EfficientNet & $\mathbf{97.97\%}$\\
    \hline    
    \end{tabular}
    \caption{Comparison of All Model Methods and Accuracy}
    \label{TableComparison}
\end{table}

From the table 1, we could observe that the EfficientNet showed the best result in modeling. We suspect that it is due to the architecture of the EfficientNet that was built on the top of MobileNet and ResNet, and it was able to capture the image features during the training.

\section{Conclusion}
This project strikes a fine balance between a research design and an implementation-centered engineering project. We believe that the key challenge to this project is identifying how to predict which charing levels that the drivers would eventually choose by the end of each trip, and which architecture would deduce the best result in predicting. From multiple trials of experiments with different input features and model architectures, we have developed that the convolutional neural network with EfficientNet that was trained on the images that were generated by the IGTD algorithm. With the current finding, we can further explore other architectures to generate artificial datasets using generative adversarial networks and then use other data analysis methods to compare the accuracy that were tested on real data and artificial data. From the paper, we have discovered that the CNN showed a promising result over machine learning models which are RF, KNN, DT.

%
\section{Future Works}
In this paper, we have used a tabular dataset we collected from the battery equipped electric vehicles to understand the charge levels of each trip. After we got the dataset, we used normalization method to convert the tabular dataset into embedding datasets. With this dataset, we created models to train convolutional neural networks, recurrent neural networks, and generative adversarial to understand how to accurately predict the charge level for each trip. From experiments on each deep learning approach, we have discovered that CNN showed the best result. However, the result did not present the best predictive model for every trip. For future works, we will implement other  model architecture such as Generative Adversarial Networks to create more artificial data to understand the dataset better to predict the charge levels. In addition, we will utilize other state-of-the-art web-based machine learning framework or services such as AWS or Microsoft Azure to train the model in more efficient ways with their servers. We expect that advanced machines would take us less time to train the model and eventually allow us try many trials for getting the best model for the experiment.

\bigskip
\section{Acknowledgement}
The team would like to express gratitude to the Institute of Transportation Studies research team at the University of California, Davis.

\bibliographystyle{abbrv}
\bibliography{refs.bib}
\end{document}